\documentclass{INTERSPEECH2023}

\usepackage{bm}
\usepackage{multirow}
\usepackage{caption}
\usepackage{subcaption}
\usepackage{booktabs} 
\usepackage[table, svgnames, dvipsnames]{xcolor}

\interspeechcameraready

\title{Bypass Temporal Classification: Weakly Supervised Automatic Speech Recognition with Imperfect Transcripts\thanks{This work was partially supported by the National Science Foundation CCRI program via Grant No 2120435.}}
\name{Dongji Gao$^1$, Matthew Wiesner$^2$, Hainan Xu$^3$, Leibny Paola Garcia$^{1,2}$, \\ 
Daniel Povey$^4$, Sanjeev Khudanpur$^{1,2}$}
\address{
  $^1$CLSP \& $^2$HLTCOE, Johns Hopkins University, USA\\
  $^3$NVIDIA, USA\\
  $^4$Xiaomi Corp., China}
\email{\{dgao5,wiesner,lgarci27,khudanpur\}@jhu.edu, hainanx@nvidia.com, dpovey@gmail.com}

\begin{document}

\maketitle
 
\begin{abstract}
This paper presents a novel algorithm for building an automatic speech recognition (ASR) model with imperfect training data. Imperfectly transcribed speech is a prevalent issue in human-annotated speech corpora, which degrades the performance of ASR models. To address this  problem, we propose Bypass Temporal Classification (BTC) as an expansion of the Connectionist Temporal Classification (CTC) criterion. BTC explicitly encodes the uncertainties associated with transcripts during training. This is accomplished by enhancing the flexibility of the training graph, which is implemented as a weighted finite-state transducer (WFST) composition. The proposed algorithm improves the robustness and accuracy of ASR systems, particularly when working with imprecisely transcribed speech corpora. Our implementation will be open-sourced.
\end{abstract}

\section{Introduction}
The quality and quantity of data are critical for training a successful ASR system. State-of-the-art end-to-end (E2E) ASR models rely heavily on large quantities of accurately annotated speech data.  
However, transcription of speech corpora by human annotators, unlike read speech, is prone to errors. Removing low-quality data can significantly reduce the amount of available data. This is especially problematic for low-resource languages and dialects, where data is scarce. 
Therefore, it is imperative to develop methods that address imperfectly labeled data for speech processing and other sequence classification tasks, to maximize the utilization of existing data.

Previous studies have focused on extracting reliable parallel speech and text from noisy data sources, such as television news broadcasts and their closed captions. While closed captions provide useful information, they are often not completely accurate, usually with a word error rate (WER) of between $10\%$ and $20\%$ compared to a careful verbatim transcript~\cite{Placeway1996CheatingWI}. To address this issue, two-step solutions have been developed: 
\begin{enumerate}
    \item Use an extra ASR model to generate hypotheses for each utterance, which is then aligned with the corresponding closed caption to identify the matching speech fragment, either at word level~\cite{Witbrock1998ImprovingAM,nguyen2004light,driesen2013lightly} or segment level~\cite{braunschweiler2010lightly,long2013improving,lanchantin2016selection,wiesner2021training,whisper}.
    \item Add the resulting speech and caption pairs to the training data to improve the ASR model.
\end{enumerate}
  This process can be repeated iteratively until adding data does not improve the ASR performance. Although these methods have provided a partial solution to the problem, their main limitation is that they are not self-contained, meaning that external ASR models are required for alignment. Furthermore, the iterative training approach can be time-consuming, which may limit its practicality in some scenarios. 

\begin{figure}[t]
\begin{minipage}[b]{.48\linewidth}
  \centering
  \centerline{\includegraphics[width=3.8cm]{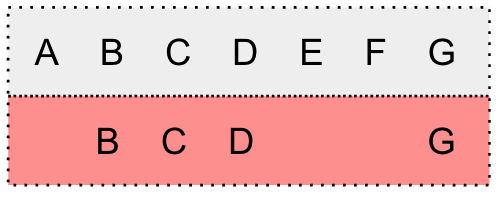}}
  \centerline{(a) Deletion (partial transcript)}
\end{minipage}
\hspace{-1mm}
\begin{minipage}[b]{.48\linewidth}
  \centering
  \centerline{\includegraphics[width=3.8cm]{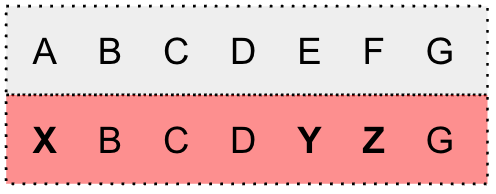}}
  \centerline{(b) Substitution}
\end{minipage}
\begin{minipage}[b]{.48\linewidth}
  \centering
  \centerline{\includegraphics[width=3.8cm]{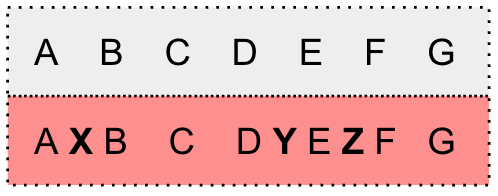}}
  \centerline{(c) Insertion}
\end{minipage}
\hspace{1mm}
\begin{minipage}[b]{.48\linewidth}
  \centering
  \centerline{\includegraphics[width=3.8cm]{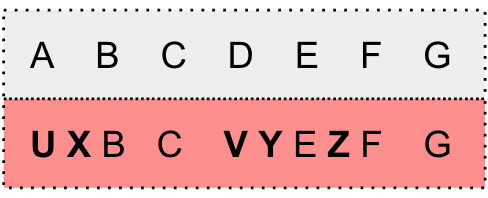}}
  \centerline{(d) Substitution and insertion}
\end{minipage}
\caption{Examples of error in the transcript. The grey box is the exact text and the red box is the imperfect text. Inaccurate words are marked in bold.}
\label{errors}
\vspace{-3mm}
\end{figure}

 \begin{figure*}[ht]
 \centering
    \begin{subfigure}[b]{0.23\textwidth}
        \centering
        \centerline{\includegraphics[width=4cm]{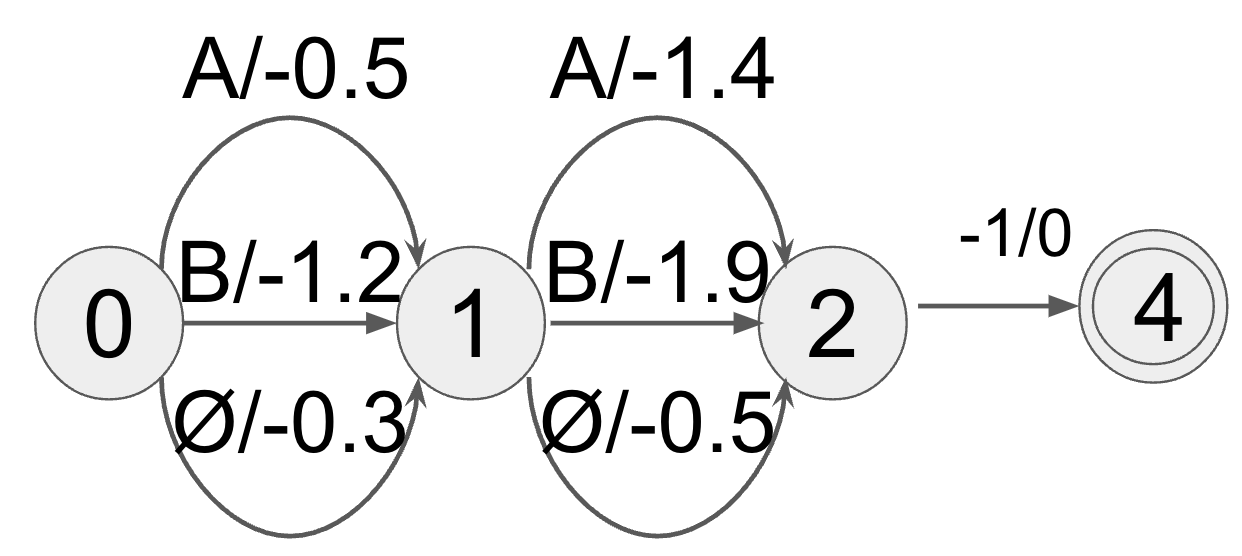}}
        \centerline{(a) Emission WFST ($E$)}
    \end{subfigure}
        \begin{subfigure}[b]{0.23\textwidth}
        \centering
        \centerline{\includegraphics[width=4cm]{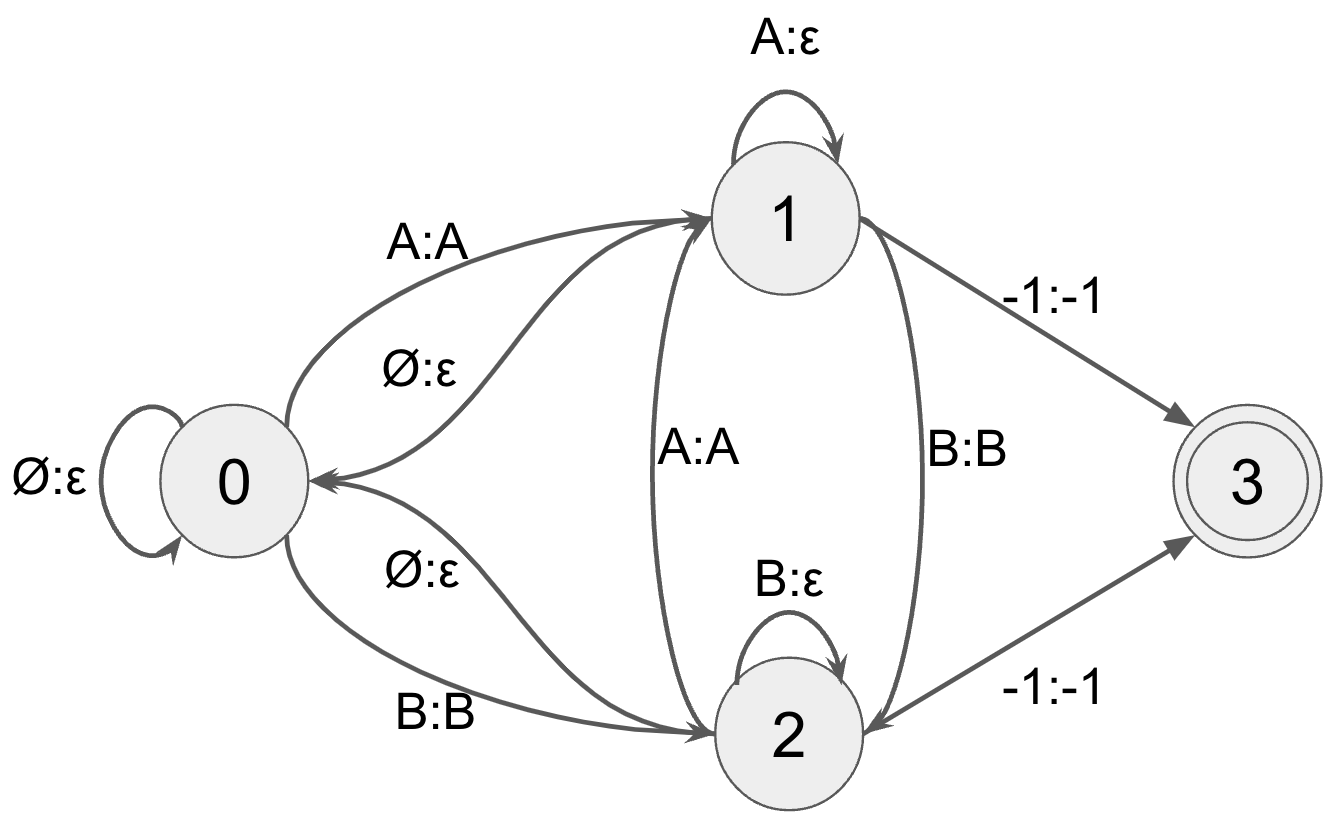}}
        \centerline{(b) Alignment WFST ($H$)}
    \end{subfigure}
    \begin{subfigure}[b]{0.23\textwidth} 
        \centering
        \centerline{\includegraphics[width=4cm]{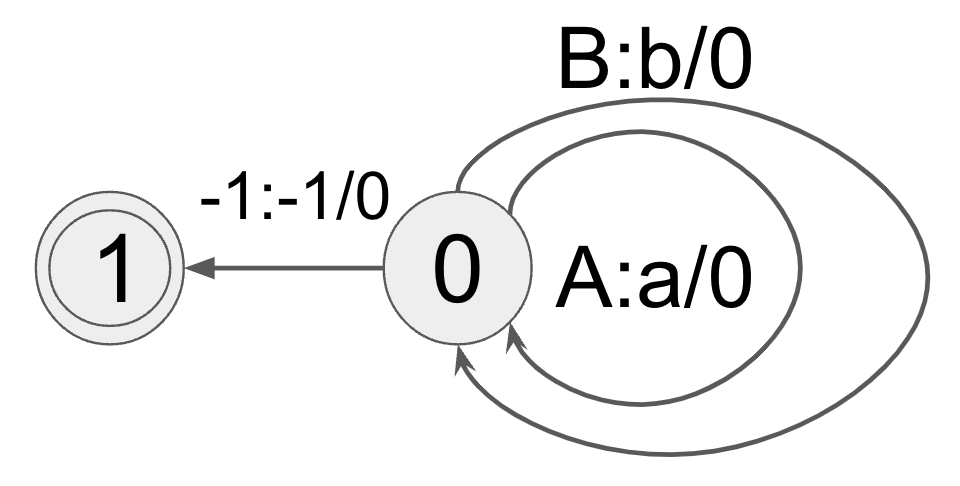}}
        \centerline{(c) Lexicon WFST ($L$)}
    \end{subfigure}
    \begin{subfigure}[b]{0.23\textwidth}
        \centering
        \centerline{\includegraphics[width=4cm]{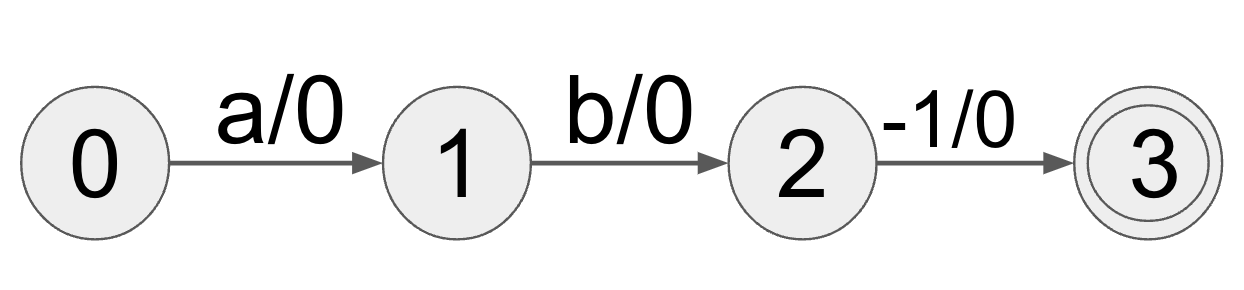}}
        \centerline{(d) Grammar WFST ($G$)}
    \end{subfigure}
 \caption{WFST topology of transcript ``a b'' with lexicon \{a:A, b:B\}. A state with $0$ is the start state. The arc with -$1$ is a special arc in k2 pointing to the final state (double circle). Each arc is associated with an input symbol, an output symbol (delimited by a colon), and weight after the slash. Arc with a single symbol indicates identical input and output. Arc without slash indicates zero weight.}
\label{ctc_topo}
\vspace{-4mm}
\end{figure*}
Alternative efforts have been proposed to directly train a model with inaccurate labels~\cite{paul2019handling}, which is commonly referred to as \emph{weakly supervised learning}. Prior research in this realm of methods has primarily concentrated on tasks involving partially labeled classification~\cite{cour2011learning,liu2014learnability}. Specifically, in ASR tasks, each spoken utterance only has partial transcriptions available (The term ``deletion'' is used in this case, as shown in Fig.~\ref{errors}(a)). To address this challenge,~\cite{wctc} proposes a novel variant of the CTC~\cite{ctc} criterion known as W-CTC. This approach enables the handling of sequences that are missing both beginning and ending labels. This work is extended by
the Star Temporal Classification (STC)~\cite{stc} to tackle the more general problem of partial labeling. The proposed STC algorithm incorporates WFSTs to explicitly manage an arbitrary number of missing labels, regardless of their location in the sequence. This study suggests that in situations where up to $70\%$ of labels are missing, the performance of STC can approach that of a supervised baseline.

Alongside this direction, we focus on the other two types of errors, i.e., substitution and insertion in transcripts, as illustrated in Fig.~\ref{errors}(b), (c), and (d).
We propose an extended CTC criterion, termed Bypass Temporal Classification (BTC), to handle substitution and insertion errors during training explicitly.
Our proposed approach can be effectively implemented under WFSTs' framework.  
We implemented WFST in the k2 toolkit\footnote{\url{https://github.com/k2-fsa/k2}}, which supports automatic differentiation for WFST to enable seamless integration between WFST and neural models. Notably, all our WFST operations can be executed on GPU, thereby significantly accelerating the training process (STC requires switching between GPU and CPU\cite{hannun2020differentiable}). We illustrate that an ASR model can be trained from scratch using BTC with imperfect transcripts, containing up to $50-70\%$ substitution or insertion errors. The model achieves acceptable ASR performance as measured by either phone error rate (PER) or word error rate (WER).

\section{Preliminaries}

\subsection{CTC}
The CTC criterion was proposed for sequence labeling and is commonly used in ASR. Given an acoustic feature sequence 
$\mathbf{x} = [x_{1},\dots,x_{T}]$ 
of length $T$, and its corresponding transcript 
$\mathbf{l} = [l_{1},\dots, l_{U}] \in \mathcal{V}^{1 \times U}$ of length $U$, where $\mathcal{V}$ is the vocabulary, and
with the constraint $U \leq T$, CTC loss is defined as
\begin{align}
    L_\text{ctc} = -\log P(\mathbf{l}|\mathbf{x}).
\end{align}
By introducing a blank token $\oslash$ and the framewise alignment sequence 
$\bm{\pi} = [\pi_{1}, \dots, \pi_{T}]$, where $\pi_{t} \in \mathcal{V} \cup \oslash$, the posterior distribution $P(\mathbf{l}|\mathbf{x})$ can be factorized as
\begin{align}
       P(\mathbf{l} | \mathbf{x}) &= \sum_{\bm{\pi} \in \mathcal{B}^{-1}(\mathbf{l})} P(\mathbf{l}, \bm{\pi} | \mathbf{x}) \\
    &= \sum_{\bm{\pi} \in \mathcal{B}^{-1}(\mathbf{l})} P(\mathbf{l} | \bm{\pi}, \mathbf{x}) P(\bm{\pi} | \mathbf{x}) \\
    &= \sum_{\bm{\pi} \in \mathcal{B}^{-1}(\mathbf{l})} P(\bm{\pi} | \mathbf{x})\label{ctc},
\end{align}
where $\mathcal{B}$ maps sequences from $\bm{\pi}$ to $\mathbf{l}$ by removing $\oslash$ and adjacent repetitions, which is why $P(\mathbf{l} | \bm{\pi}, \mathbf{x}) \equiv 1$ for every $\bm{\pi} \in \mathcal{B}^{-1}(\mathbf{l})$.  $P(\bm{\pi} | \mathbf{x})$ may be further factorized as
\begin{equation}
\begin{aligned}
    P(\bm{\pi} | \mathbf{x}) &= \prod_{t=1}^{T} P(\pi_{t} | \pi_{1},\dots,\pi_{t-1},\mathbf{x}) \\
    &= \prod_{t=1}^{T} P(\pi_{t} | \mathbf{x}),
\end{aligned}
\end{equation}
by {\em assuming} that $\bm\pi$ is {\em conditionally i.i.d.} given the sequence $\mathbf{x}$.
\vspace{-1mm}

\subsection{WFST}\label{wfst}
The marginalization over token sequences represented by $\mathcal{B}^{-1}\left(\mathbf{l}\right)$ can be efficiently computed using WFSTs. A WFST maps input symbol sequences to output symbol sequences and assigns a weight to each transition in the transducer. 
The weight of the mapping is often set to be the conditional probability of the output sequence given the input. WFSTs have found extensive application in the field of ASR to model the probability of decoding unit sequences~\cite{wfst,Povey2011TheKS,Povey2012GeneratingEL}. This can be attributed to the compactness of its representation of the model architecture~\cite{miao2015eesen,laptev2022ctc}.

Two WFSTs can be composed to cascade mapping operations. Given  
a WFST $F_{1}$ that maps $\mathbf{a}$ to $\mathbf{b}$ with weight $w_{1}$, and a WFST $F_{2}$ that maps $\mathbf{b}$ to $\mathbf{c}$ with weight $w_{2}$. The composed WFST, denoted $F_{1}$ $\circ$ $F_{2}$, maps $\mathbf{a}$ to $\mathbf{c}$. Its weight can be either $\max(w_1, w_2)$ (tropical semiring) or $\text{log-sum-exp}(w_1,w_2)$ (log semiring). 
\vspace{-2mm}
\begin{figure}[h]
  \centering
  \includegraphics[width=0.8\linewidth]{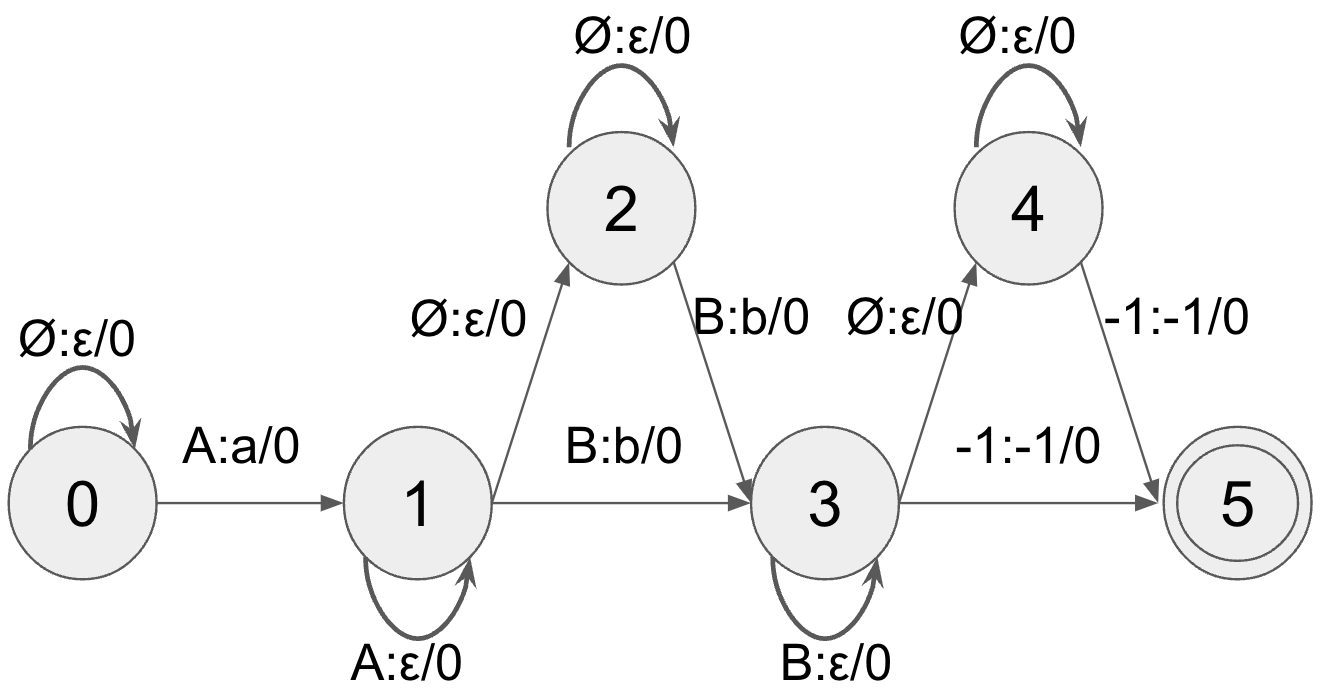}
\vspace{-2mm}
  \caption{CTC graph of ``a b''. $\oslash$ is the blank token and $\epsilon$ represents no output symbol in WFST transitions. The reader can refer to Fig.~\ref{ctc_topo} for details of WFST representation.}
  \label{fig:hlg_ctc}
\end{figure}

The composition of the alignment WFST ($H$), with a WFST representation of a lexicon ($L$) and transcript ($G(\mathbf{l})$) and inverting the result, returns a transducer that maps from transcripts to a graph that represents the space of all possible CTC paths associated with the transcript. Marginalization is enabled via the application of the forward algorithm on the resulting lattice of paths. We show that Eq.~\ref{ctc} may be represented using a composition of WFSTs:
\begin{equation}
    P(\mathbf{l} | \bm{\pi}) \,\,\,\,\,\,= 
    \sum_{\bm{\pi} \in (H \circ L \circ G(\mathbf{l}))^{-1}}
    \underbrace{P(\bm{\pi} | \mathbf{x})}_{
        \textstyle
            \begin{gathered}
                E 
            \end{gathered}
            },
\end{equation}
where the superscript $-1$ denotes WFST inversion;
\begin{itemize}
    \item $E$ is the emission WFSA representing $P(\bm{\pi} | \mathbf{x})$, as shown in Fig.~\ref{ctc_topo}(a). For state $t$, the weights on its arcs are the log-probabilities on the set of {\em decoding units} (e.g., characters or phones or sub-words or words, and $\oslash$) at time frame $t$;
    \item $H$ is the CTC alignment WFST. It removes $\oslash$ and adjacent repeated decoding units;
    \item $L$ is the lexicon. It maps sequences of decoding units to words. It is discarded if the decoding units are words; 
    \item $G(\mathbf{l})$ is the grammar or language model of $\mathbf{l}$, assumed represent-able as a WFST, as shown in Fig.~\ref{ctc_topo}(d).
\end{itemize}
Therefore, $\log P(\mathbf{l}|\mathbf{x})$ equals to the total weight of $E \circ H \circ L \circ G(\mathbf{l})$
under the log-semiring. We refer to $H \circ L \circ G(\mathbf{l})$ as {\em the graph} of $\mathbf{l}$, and illustrate it with an example in Fig.~\ref{fig:hlg_ctc}.

\section{Method}

In the training process of CTC models, for each training example  ($\mathbf{x}, \mathbf{l}$), a graph is constructed  based on $\mathbf{l}$. The presence of substitution and insertion errors in $\mathbf{l}$ can lead to erroneous graph construction, causing misalignment between certain segments of $\mathbf{x}$ and incorrect tokens. The consequential misalignment can adversely affect the ASR model through back-propagation. 

\begin{figure}[t]
  \centering
  \includegraphics[width=0.6\linewidth]{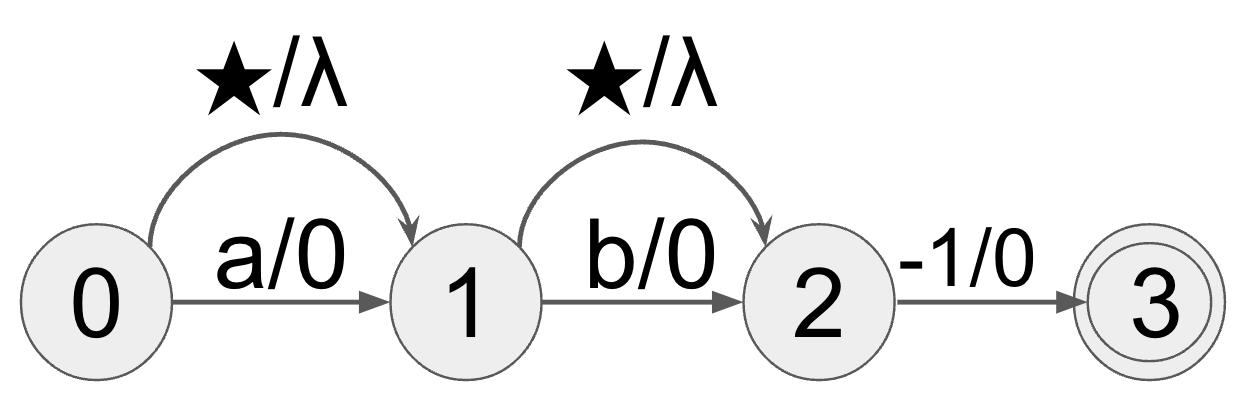}
  \vspace{-2mm}
  \caption{BTC Grammar WFST (G) of transcript ``a b''. Each label has parallel arcs added with input symbol $\star$ and weight $\lambda$.}
  \label{fig:btc grammar graph}
\vspace{-4mm}
\end{figure}

\begin{figure}[h]
  \centering
  \includegraphics[width=0.9\linewidth]{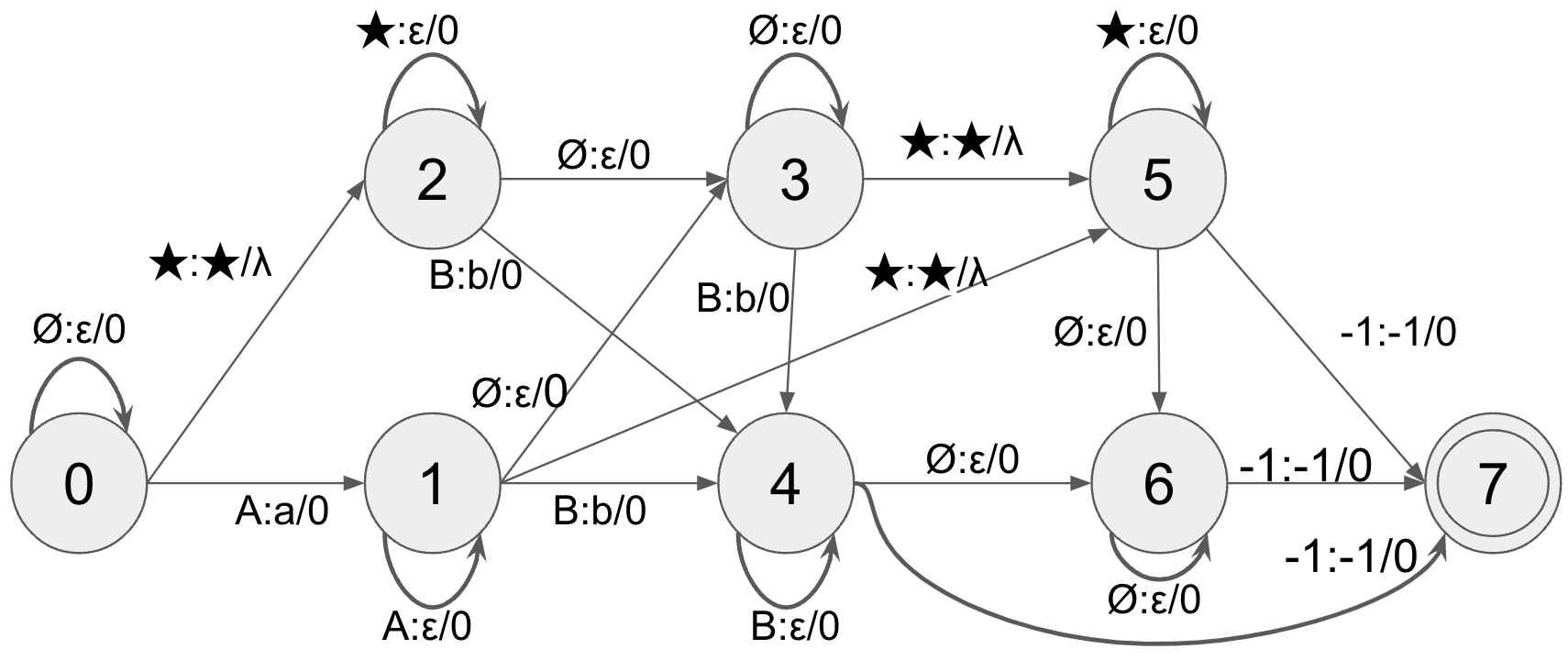}
  \vspace{-2mm}
  \caption{Topology of the BTC graph for ``a b'' after composition with the CTC alignment transducer $H$.}
  \label{fig:hlg_btc}
\vspace{-2mm}
\end{figure}

In the CTC framework, $P(\pi_t | \bm{x})$ represents the underlying model's estimate of ``which decoding unit best represents this frame.'' Setting $\pi_t=\oslash$ represents a continuation-without-explicit-repetition of the most recent non-blank symbol in $\bm{\pi}$, allowing for {\em uncertainty in the temporal span} of the words in $\mathbf{l}$.  But there is no uncertainty in $\mathbf{l}={\cal B}(\bm\pi)$.
In BTC, we introduce a special ``wildcard'' token $\star$ to model {\em uncertainty in the transcript} $\mathbf{l}$. The $\star$ arcs are added to $G(\mathbf{l})$ in parallel to words in the transcript, as shown in Fig.~\ref{fig:btc grammar graph}, and to $L$ as an identity mapping (i.e. with no subword decomposition).  The alignment transducer $H$ treats $\star$ as a standard decoding unit, resulting in a graph of the kind shown in Fig.~\ref{fig:hlg_btc}.  This topology is capable of modeling insertions and substitution errors in $\mathbf{l}$:
\begin{enumerate}
    \item Substitution: since there is a $\star$ arc parallel to arc of an incorrect word $l$, $P(\pi_t | \bm{x})$ can bypass $l$ and assign the acoustics of the (unknowable) correct word to $\star$ with high probability, thereby avoid corruption of its internal representation of $l$: $\star$ acts like a garbage collector.
    \item Insertion: when $\mathbf{l}$ contains more words than were spoken, $P(\pi_t | \bm{x})$ can again bypass an inserted word $l$, assigning a {\em minimally necessary} number of frames to the $\star$ arc in parallel to $l$, and continue.  This again avoids corrupting the internal representation of $l$, albeit by ``stealing'' a minimal number of frames from the adjacent words to assign to $\star$.
\end{enumerate}

Training naively on the BTC grammar with random initialization, however, leads to a degenerate solution: map all acoustic features to $\star$ with high probability. In order to avoid this solution, we introduce a penalty, i.e. a negative weight denoted by $\lambda$, to all $\star$ arcs.  We then investigate a scheduling algorithm that adjusts this penalty during training: it is initially set to a large value to encourage the model to start learning the correspondence between acoustics and the imperfect transcripts. As the training progresses, and the models understanding of this relationship begins to take form, we gradually decrease the penalty. Specifically, for the $i$-th epoch, the penalty is set as
\begin{equation}
    \lambda_{i} = \beta * \tau^{i},
\end{equation}
where $\beta$ is the initial (high) penalty that decays geometrically by a factor $\tau \in (0, 1)$. We determine $\beta$ and $\tau$ empirically.
\vspace{-2mm}

\section{Experimental Setup}
\subsection{Datasets}

We use two standard datasets for our experiments.\vspace{1mm}

\noindent \textbf{TIMIT}~\cite{timit} is comprised of 6300 sentences, totaling 5.4 hours of recorded read speech. Each sentence has been phonetically transcribed. We adhere to the standard training, development, and test partitions, consisting of 3696, 400, and 192 sentences, respectively. \vspace{1mm}

\noindent \textbf{LibriSpeech}~\cite{librispeech} is comprised of 960 hours of read speech for training, further divided into three subsets of 100, 360 and 500 hours of clean speech and ``other'' speech. Two development subsets and two test subsets, each consisting of 5 hours of speech, are also included in the data set.
\vspace{-1mm}

\subsection{Imperfect transcripts generation}
In our study, we generate synthetically erroneous transcripts from the provided transcripts, which we assume to be perfect, for three scenarios.
\begin{itemize}
    \item For substitution, each token in the transcript is replaced by a random token with $p_{\text{sub}} \in [0.1, 0.3, 0.5, 0.7]$. 
\item For insertion, a random token is inserted with $p_{\text{ins}} \in [0.1, 0.3, 0.5, 0.7]$ between any two tokens in the transcript.
\item For substitution$+$insertion, insertion is followed by substitution with $p_{\text{sub}} \in [0.05, 0.15, 0.25, 0.35]$ and $p_{\text{ins}} \in [0.05, 0.15, 0.25, 0.35]$.
\end{itemize}
Different choices of $p_{\text{sub}}$ and $p_{\text{ins}}$ are studied and compared. 

\begin{table*}[t]
\centering
\caption{WER ($\%$) on LibriSpeech test-clean and test-other dataset. We compared CTC (highlighted in grey) and BTC in three scenarios: substitution-only, insertion-only, and substitution and insertion. We measure WER with and without LM (delimited by a slash). ``-'' indicates that the model does not converge.}
\vspace{-2mm}
\label{librispeech_wer}
\begin{tabular}{llllllllll}
\bottomrule
\multirow{2}{*}{Error}   & \multicolumn{1}{c}{\multirow{2}{*}{Criterion}} & \multicolumn{8}{c}{$p_\text{sub}, p_\text{ins}, p_\text{sub+ins} $}                                                                        \\ \cline{3-10} 
   & \multicolumn{1}{c}{}           & \multicolumn{2}{c}{0.1} & \multicolumn{2}{c}{0.3} & \multicolumn{2}{c}{0.5} & \multicolumn{2}{c}{0.7} \\ \cline{3-10} 
                         &                                & clean      & other      & clean      & other      & clean      & other      & clean      & other      \\ \bottomrule 
\rowcolor{Gainsboro!60}
\multirow{2}{*}{sub}     & CTC                            &6.9/15.4 &15.7/29.2 & 12.9/20.4 &24.5/36.5 &36.1/43.2 &54.9/60.5     &-/-   &-/-            \\
                         & BTC                            &6.1/14.7 &14.5/29.0 & 6.6/17.5 & 15.6/33.3  &7.4/19.8 &18.0/36.5   &-/-  &-/-    \\ \bottomrule
\rowcolor{Gainsboro!60}
\multirow{2}{*}{ins}     & CTC                             &6.9/16.6 &15.8/31.0 &19.5/29.3 &30.0/44.3    &-/-         &-/-        &-/-          &-/-            \\
                         & BTC                             &5.5/12.0 &14.1/24.0  &5.6/12.1  &14.1/24.1   &5.6/12.1  &14.2/24.4 & 6.0/12.7 & 14.8/24.6    \\ \hline
\rowcolor{Gainsboro!60}
\multirow{2}{*}{sub+ins} & CTC                          &7.3/17.3 &16.2/31.9 &19.5/25.1 &31.9/41.2   &45.6/45.4 &59.94/61.2            & -/-           & -/-           \\
                         & BTC                          &5.8/13.9 &13.8/27.8 &6.2/15.6 &14.1/29.8 &6.6/16.2 &14.8/30.8 &6.9/16.9 &15.6/32.3            \\ \bottomrule
\end{tabular}
\vspace{-3mm}
\end{table*}

\vspace{-1mm}
\subsection{Acoustic feature}
We utilize a pre-trained wav2vec 2.0 model~\cite{wav2vec2} (wav2vec2-base) to extract 768-dimensional acoustic features. This self-supervised model contains 12 transformer blocks with 8 attention heads~\cite{transformer}. The acoustic features are extracted using the S3PRL toolkit.

\subsection{Model architecture}
We investigate two distinct model architectures utilizing the BTC criterion: 

\noindent \textbf{TDNN-LSTM}~\cite{Sak2014LongSM,Peddinti2018LowLA}: It combines time-delay neural networks (TDNN) and long short-term memory(LSTM)~\cite{Hochreiter1997LongSM} networks. The model consists of 3 TDNN layers followed by the LSTM layer, and a linear decoder layer.
This hybrid architecture can effectively capture both local and global dependencies in speech signals. 
\vspace{1mm}

\noindent \textbf{Conformer}~\cite{conformer}: It employs a combination of convolutional neural networks (CNNs)~\cite{LeCun1998ConvolutionalNF} and self-attention mechanisms to model speech signals effectively. The Conformer encoder in our study consists of 12 layers, with each layer comprising a conformer block that includes a convolution module stacked after the self-attention module. The decoder is simply a linear layer.
\vspace{-2mm}

\section{Results}

\subsection{TIMIT}
We first evaluate BTC on the TIMIT dataset to gain insight into BTC as well as tuning hyper-parameters, specifically the initial penalty and penalty decay factor. The TDNN-LSTM model is utilized, with phone as the decoding unit.  
\vspace{-2mm}
\begin{table}[h]
\centering
\caption{PER ($\%$) on TIMIT test set with substitution error. CTC is highlighted in grey.}

\label{timit_per}
\begin{tabular}{llllll}
\bottomrule
\multicolumn{1}{c}{\multirow{2}{*}{Error}} & \multicolumn{1}{c}{\multirow{2}{*}{Criterion}} & \multicolumn{4}{c}{$p_\text{sub}$,$p_\text{ins}$,$p_\text{sub+ins}$} \\ \cline{3-6} 
\multicolumn{1}{c}{}                       & \multicolumn{1}{c}{}                           & 0.1    & 0.3   & 0.5   & 0.7   \\ \bottomrule
\rowcolor{Gainsboro!60}
\multirow{2}{*}{sub}                       & CTC                                            & 28.3  & 40.9 & 49.2 & 63.1 \\
                                            & BTC                                           & 13.1  & 16.8 & 17.2 & 21.4 \\ \bottomrule
\rowcolor{Gainsboro!60}
\multirow{2}{*}{ins}                       & CTC                                            & 16.1  & 18.1 & 28.8 & 32.8      \\
                                           & BTC                                            & 13.6  & 14.2 & 14.3 & 14.7 \\ \bottomrule
\rowcolor{Gainsboro!60}
\multirow{2}{*}{sub+ins}                   & CTC                                            & 20.3  & 27.7 & 31.2  & 42.6      \\
                                           & BTC                                            & 13.4  & 14.0 & 15.9 & 21.4 \\ \bottomrule
\end{tabular}
\end{table}

We establish a CTC baseline and compare the phone error rate (PER) between BTC and CTC. 
Our findings, presented in Table~\ref{timit_per}, demonstrate that CTC's performance degrades as the error rate increases. Even with a relatively low mistranscription rate of $10\%$, the PER increases from 13.51 to 28.33, indicating a near-doubling of the error rate. The PER can reach as high as 63.13 under more severe mistranscription conditions ($70\%$ substitution error rate). 
In contrast, our evaluation reveals that the BTC shows high robustness across varying error scenarios. Even when faced with a challenging scenario involving both substitution and insertion errors, BTC achieved a phone error rate (PER) of approximately $20$, despite $70\%$ of transcripts being incorrect.
\vspace{-1mm}

\subsection{LibriSpeech}

For LibriSpeech, we use clean 100 hours set as the training set.
As mentioned in Section~\ref{wfst}, the use of a lexicon WFST $L$ converts decoding units to words, thereby enabling the measurement of WER.
\vspace{-1mm}
\subsubsection{Model and decoding unit selection}

Given the similarity in trend between substitution and insertion, we focus on comparing TDNN-LSTM and Conformer models in the substitution case. Phones are utilized as the decoding unit, and the findings presented in Table~\ref{tdnn_conformer} demonstrate the consistent superiority of the Conformer model over TDNN-LSTM in terms of word error rate (WER) across different substitution error rates. So we choose the Conformer model for the following experiments.

\begin{table}[h]
\centering
\caption{WER ($\%$) of TDNN-LSTM and Conformer with substitution error. ``-'' indicates that the model does not converge.}
\label{tdnn_conformer}
\vspace{-2mm}
\begin{tabular}{lllll}
\hline
\multicolumn{1}{c}{\multirow{2}{*}{Architecture}} & \multicolumn{4}{c}{$p_\text{sub}$} \\ \cline{2-5} 
\multicolumn{1}{c}{}                              & 0.1       & 0.3       & 0.5       & 0.7     \\ \hline
TDNN-LSTM                                         & 6.8       & 7.1      & 8.6      & -       \\
Conformer                                         & 6.1      & 6.6      & 7.4      & -        \\ \hline
\end{tabular}
\vspace{-3mm}
\end{table}

We then compare the effectiveness of BPE subwords and phones for decoding using Conformer architecture. We conduct the analysis of phones against BPE with vocabulary sizes of 100 and 500. Our results in Table~\ref{phone_bpe} indicate that phones outperform BPE in terms of performance and robustness. Notably, BPE fails to converge at a substitution error rate of 0.5, even with a comparable vocabulary size of phones (100 versus 71). This is attributed to the ability of phones to capture acoustic feature patterns.
\vspace{-1mm}
\begin{table}[h]
\centering
\caption{WER ($\%$) of phone and BPE with substitution error. ``-'' indicates that the model does not converge.}
\label{phone_bpe}
\vspace{-2mm}
\begin{tabular}{llllll}
\hline
\multicolumn{1}{c}{\multirow{2}{*}{Unit}} & \multicolumn{1}{c}{\multirow{2}{*}{Vocab size}} & \multicolumn{4}{c}{$p_\text{sub}$} \\ \cline{3-6} 
\multicolumn{1}{c}{}                               & \multicolumn{1}{c}{}                            & 0.1       & 0.3       & 0.5       & 0.7     \\ \hline
BPE                                                & 500                                             & 6.5      & 7.1      & -         & -       \\
                                                   & 100                                             & 6.3          & 6.7           & -          & -        \\ \hline
phone                                            & 71                                              & 6.1      & 6.6      & 7.4     & -       \\ \hline
\end{tabular}
\vspace{-5mm}
\end{table}

\subsubsection{Results}

We show the result with and without integrating a tri-gram language model (LM) in Table~\ref{librispeech_wer}. The performance of BTC is consistent with that observed in the TIMIT dataset. The ASR model failed to converge when subject to a substitution error rate of $70\%$. Other than that, the system trained using BTC still remains satisfactory, with minor degradation. Moreover, for error-free transcripts, BTC does not hurt the performance compared with CTC: $5.5$ vs.\ $5.6$ on test-clean and $14.4$ vs.\ $14.5$ on test-other.
\vspace{-1mm}

\section{Conclusion and future work}

This research introduces BTC as a promising approach for weakly supervised ASR. BTC, a variant of the CTC algorithm, enables ASR model training from scratch using an imperfect transcript that contains substitution and insertion errors. The implementation of the algorithm is efficient within the WFST framework, and all operations can be fully performed on GPU to expedite the training process. Our experiments on the TIMIT and LibriSpeech datasets demonstrate that the BTC approach can effectively train an ASR model without much degradation occurring when $50\%$ to $70\%$ of the transcripts contain errors. 

Moving forward, our goal is to integrate BTC with STC to address all three types of errors (deletion, substitution, and insertion) within a unified framework for weakly supervised tasks.
\clearpage
\bibliographystyle{IEEEtran}
\bibliography{mybib}

\end{document}